\documentclass[sigconf]{acmart}

\usepackage{balance}
\usepackage{multirow}
\usepackage{booktabs}
\usepackage{amsmath}
\usepackage{bm}
\usepackage{enumitem}
\begin{document}

\fancyhead{}

\title{Quantifying and Alleviating the Language Prior Problem in \\ Visual Question Answering}

\copyrightyear{2019}
\acmYear{2019}
\setcopyright{acmcopyright}
\acmConference[SIGIR '19]{Proceedings of the 42nd International ACM SIGIR Conference on Research and Development in Information Retrieval}{July 21--25, 2019}{Paris, France}
\acmBooktitle{Proceedings of the 42nd International ACM SIGIR Conference on Research and Development in Information Retrieval (SIGIR '19), July 21--25, 2019, Paris, France}
\acmPrice{15.00}
\acmDOI{10.1145/3331184.3331186}
\acmISBN{978-1-4503-6172-9/19/07}

\author{\mbox{Yangyang Guo$^\dag$, Zhiyong Cheng$^{\S}$, Liqiang Nie$^{\dag}$,  Yibing Liu$^\dag$, Yinglong Wang$^\S$, Mohan Kankanhalli$^\ddag$}}
\affiliation{\institution{$^\dag$School of Computer Science and Technology, Shandong University}}
\affiliation{\institution{$^\S$Shandong Computer Science Center (National Supercomputer Center in Jinan), \\ Qilu University of Technology (Shandong Academy of Sciences)}}
\affiliation{\institution{$^\ddag$School of Computing, National University of Singapore}}
\email{{guoyang.eric, jason.zy.cheng, nieliqiang, lyibing112}@gmail.com, wangyl@sdas.org, mohan@comp.nus.edu.sg}
%


\thanks{Corresponding Author: Zhiyong Cheng and Liqiang Nie.}

\begin{abstract}
Benefiting from the advancement of computer vision, natural language processing and information retrieval techniques, visual question answering (VQA), which aims to answer questions about an image or a video, has received lots of attentions over the past few years. Although some progress has been achieved so far, several studies have pointed out that current VQA models are heavily affected by the \emph{language prior problem}, which means they tend to answer questions based on the co-occurrence patterns of question keywords (e.g., \emph{how many}) and answers (e.g., \emph{2}) instead of understanding  images and questions. Existing methods attempt to solve this problem by either \emph{balancing the biased datasets} or \emph{forcing models to better understand images}. However, only marginal effects and even performance  deterioration are observed for the first and second solution, respectively. In addition, another important issue is the lack of measurement to quantitatively measure the extent of the language prior effect, which severely hinders the advancement of related techniques.

In this paper, we make contributions to solve the above problems from two perspectives. Firstly, we design a metric to quantitatively measure the language prior effect of VQA models. The proposed metric has been demonstrated to be effective in our empirical studies. Secondly, we propose a regularization method (i.e., score regularization module) to enhance current VQA models by alleviating the language prior problem as well as boosting the backbone model performance. The proposed score regularization module adopts a pair-wise learning strategy, which makes the VQA models answer the question based on the reasoning of the image (upon this question) instead of basing on question-answer patterns observed in the biased training set. The score regularization module is flexible to be integrated into various VQA models. We conducted extensive experiments over two popular VQA datasets (i.e., VQA 1.0 and VQA 2.0) and integrated the score regularization module into three state-of-the-art VQA models. Experimental results show that the score regularization module can not only effectively reduce the language prior problem of these VQA models but also consistently improve their question answering accuracy.
\end{abstract}



\maketitle

\section{Introduction}
Question Answering (QA) has been long recognized as a challenging information retrieval task. At the beginning, it focuses only on the text domain. With the great progress of natural language processing (NLP), computer vision (CV) and information retrieval (IR), a new `AI-complete' task, namely visual question answering (VQA), has become an emerging interdisciplinary research field over the past few years. VQA aims to accurately answer natural language questions about a given image or a video, bringing bright prospect in various applications including medical assistance and human-machine interaction. Recently, several benchmark datasets have been constructed to facilitate this task~\cite{vqa1, genome, multi-world, visual7w}, followed by a number of devised deep models~\cite{vqa1, multi-world, hierarchical, san, nmn, askyour}.

Although these methods have achieved state-of-the-art performance over their contemporary baselines, many studies point out that today's VQA models are yet heavily driven by superficial correlations between questions and answers in the training data and lack of sufficient visual understanding~\cite{overcomepriorcvpr, overcomepriornips, vqadatasets}. As a consequence, it turns out that the carefully designed VQA models actually provide answers based upon the first few words in questions and can frequently yield not bad performance. Taking the VQA 1.0 training set~\cite{vqa1} as an example, \emph{2} accounts for 31\% of the answers to questions initiating with \emph{how many}. This leads to VQA models overwhelmingly replying to `how many ...' questions with \emph{2} without truly understanding the given images when testing. The problem that \emph{the predicted answers are strongly driven by the answer set from the same question type\footnote{Questions initiate with the same words.} in the training set} is the so called \emph{language prior problem}~\cite{vqa2, survey, overcomepriorcvpr} that many VQA models confront with.

It is not hard to understand the reason of the language prior problem, however, this problem is non-trivial to deal with. One reason for this unsatisfactory behavior is the fundamentally problematic nature of independent and identically distributed (i.e., IID) train-test splits in the presence of strong priors. Accordingly, it is hard to distinguish a well-performed model between making progress towards the goal of understanding images correctly and only exploiting language priors to achieve high accuracy~\cite{overcomepriorcvpr}. Moreover, tackling with the language prior problem without deteriorating the model performance poses another challenge.

With the realization of the language prior problem in VQA, researchers have devoted great efforts to solving or somehow alleviating the problem and developed a set of approaches. Existing approaches can be broadly classified into two directions: 1) making the datasets less biased; and 2) making the model answer questions by analyzing the image contents. In the first direction, researchers in~\cite{yinyang, vqa2} tried to balance the existing VQA 1.0 dataset by adding complementary entries and built the VQA 2.0 dataset~\cite{vqa2}. More concretely, for each <image, question, answer> triplet, another triplet with a visually similar image but a different answer is collected to elevate the role of images in VQA. However, even with this balance, there still exists significant bias in the augmented VQA 2.0 datasets. For instance, \emph{2} still accounts for 27\% of the question type \emph{how many} in the training set of the VQA 2.0 dataset. Instead of amending the datasets, Johnson et al.~\cite{clevr} designed a diagnostic 3D shape dataset to control the question-conditional bias via rejection sampling within families of related questions. Since they dealt with the problem from the perspective of datasets and attempted to circumvent the inherent deficiency in traditional biased datasets, the language prior problem of previous methods still remains unsettled.

In contrast, researchers in the second direction make efforts to design mechanisms to make the VQA models avoid the language prior problem. Approaches in this direction can be directly used in the biased datasets and thus are more generalizable. For example, the method in~\cite{overcomepriorcvpr} explicitly disentangles the recognition of visual concepts present in the image from the answer prediction for a given question. And more recently, Ramakrishnan et al.~\cite{overcomepriornips} treated the training as an adversarial game between the VQA model and the QA model (eliminating images from the current triplet) to reduce the impact of language biases. Both methods are built upon the widely used VQA model Stacked Attention Networks (SAN)~\cite{san}. Nevertheless, performance deterioration is observed for both methods as compared to the backbone model SAN. We argue that a better regularization can not only alleviate the language prior problem but also improve the model performance.

Another important issue is the lack of proper evaluation metrics to measure the extent of language prior effect of VQA models. Although the language prior problem has been pointed out by various previous studies~\cite{vqa2, yinyang, vqadatasets, analyzing, clevr} and many approaches have been proposed to deal with this problem~\cite{overcomepriorcvpr, overcomepriornips}, few efforts have been devoted into how to numerically quantify the language prior effect. As discussed, it is hard to distinguish whether the model really understands the question and image contents before answering the question or it just simply discovers some patterns between question words and answers. Besides, it is also difficult to evaluate how well a newly designed model solves the language prior problem.

In order to tackle the aforementioned limitations of the previous approaches and the lack of language prior measurement, in this paper, we establish a formal quantitative metric to measure the extent of language prior effect (called LP score) and design a generalized regularization method to alleviate the language prior problem in VQA. On the one hand, the proposed LP score evaluates the language prior effect by taking into account both the training dataset bias and model deficiency. In this way, the LP score can measure the language prior effect quantitatively and guide further studies on alleviating the language prior problem. On the other hand, our proposed regularization method leverages a \emph{score regularization module} to force backbone models to better reason the image contents before predicting answers. More specifically, the score regularization module is added to the backbone models before the final answer prediction layer. This is to guarantee that the VQA model answers questions by understanding questions and corresponding image contents instead of simply analyzing the co-occurrence patterns of question key words (e.g., \emph{how many}) and the answer (e.g., \emph{2}). To achieve the goal, the inputs to the score regularization module are from two streams:  \emph{fused question-image feature with the embedding feature of the true answer} and \emph{question feature with the embedding feature of the true answer}; and then the score regularization module computes two scores and adopts a pair-wise learning scheme for training. Different from the multi-step learning as adopted in ~\cite{overcomepriorcvpr, overcomepriornips}, we train the proposed regularizer with the backbone model in an end-to-end multi-task learning scheme. Moreover, our proposed regularization method can be applied to most of the existing VQA models on the biased datasets.

To verify the effectiveness of our proposed regularization method, we conducted extensive experiments on two most popular datasets VQA 1.0~\cite{vqa1} and VQA 2.0~\cite{vqa2}. Moreover, we added the proposed regularization module three state-of-the-art models. Experimental results demonstrate that our proposed methods can yield better performance as compared to the corresponding backbone models, and thus achieve state-of-the-art performance.

In summary, our main contributions in this paper are threefold:
\begin{itemize}[align=left,style=nextline,leftmargin=*,labelsep=\parindent,font=\normalfont]
  \item To the best of our knowledge, we are the first to study the lack of language prior measurement and emphasize its importance on facilitating the advancement of related techniques. We further design an evaluation metric to quantitatively measure the extent of  language prior effect of VQA models.
  \item We propose a regularization method on VQA models by forcing the backbone model to better reason visual contents before the final answer prediction. The proposed regularizer can help reduce the language prior effect as well as boost the model performance. It is excepted that our method can be extended to other visual-language reasoning tasks which also suffer from the language prior problem, e.g., image captioning.
  \item We conducted extensive comparative experiments on two publicly available datasets to validate the effectiveness of the proposed regularization method and the feasibility of the proposed evaluation metric. Moreover, we have released the codes and setting to facilitate future research in this direction\footnote{\href{https://github.com/guoyang9/vqa-prior}{https://github.com/guoyang9/vqa-prior.}}.
\end{itemize} 
\section{Language Prior Measurement} \label{measure}
%

\subsection{Observations}
\begin{figure}
  \centering
  \includegraphics[width=\linewidth]{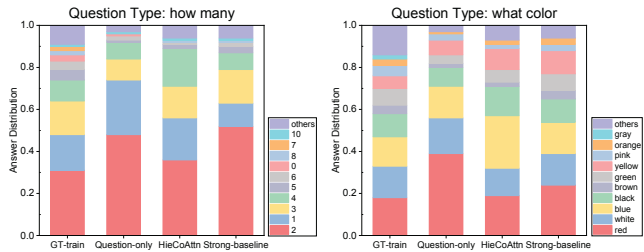}
  \caption{Answer distribution of two question types in the VQA 1.0 dataset.}\label{fig:prior1}
  \vspace{-1em}
\end{figure}

Before elaborating the conception of our language prior measurement, let us go through some examples to show the language prior problem intuitively. Figure~\ref{fig:prior1} shows the answer distributions in the VQA 1.0 dataset~\cite{vqa1} of two question types: \emph{how many} and \emph{what color}\footnote{These two types of questions take about 20\% of the questions in the whole dataset.}. For both sub figures, the leftmost bar represents the ground truth answer distribution (i.e., GT-train) in the training set. For example, the 31\% answers to the question type \emph{how many} are \emph{2}. And the right ones are the distributions of the predicted wrong answers of three baselines in the validation set. The Question-only~\cite{vqa1, vqa2} is the model trained without reasoning the image, and it is for sure that this baseline would arise the language prior problem. The other two HieCoAttn~\cite{hierarchical} and Strong-baseline~\cite{strongbaseline} are the state-of-the-art models on the VQA 1.0 dataset. For example, if the true answer for a given question of the question type \emph{how many} is \emph{6}, and the predicted answer is \emph{2}, then it is counted as a predicted wrong answer of \emph{2} under the question type \emph{how many}. Without dataset bias or language prior, the predicted wrong answers from VQA models should be more diverse (\emph{5}, \emph{apple}, \emph{on the left}, etc) or roughly follow a uniform distribution over all answers instead of being proportional to the answer distribution in the training set. For example, a large portion of answers from \emph{how many} questions are mispredicted \emph{2}, \emph{1} and \emph{3}, which are also the most frequent true answers in the training dataset (as shown in Figure~\ref{fig:prior1}). This indicates that VQA models tend to provide answers according to the patterns observed between question types and answers in the training set rather than reasoning images upon the current question.

\begin{figure}
  \centering
  \includegraphics[width=\linewidth]{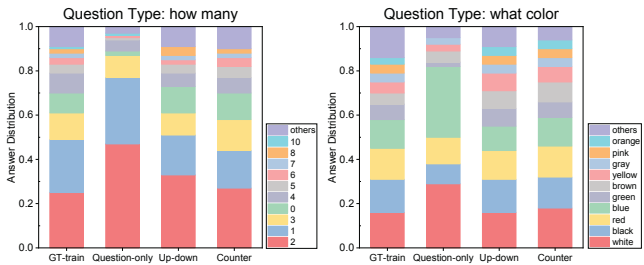}
  \caption{Answer distribution of two question types in the VQA 2.0 dataset.}\label{fig:prior2}
  \vspace{-1em}
\end{figure}

In the recent work, Goyal et al.~\cite{vqa2} attempted to deal with the language prior problem of the popular VQA 1.0 dataset by collecting <image, question, answer> triplets to construct the VQA 2.0 dataset. Instead of associating each question with just one image as in the VQA 1.0 dataset, the VQA 2.0 dataset assigns a pair of similar images with different answers to the same question. However, as shown in Figure~\ref{fig:prior2}, the language prior problem still exists. The predicted wrong answer distributions from the state-of-the-art models Up-Down~\cite{updown} and Counter~\cite{counter} are still high biased.

The phenomenon arises from the two aspects: 1) training dataset bias. It is common that some answers are more frequently to be replied to a certain question type. For example, \emph{red}, \emph{white}, \emph{blue} and \emph{black} are the most frequent colors in daily life, constituting the most frequent answers to the question type \emph{what color}. Moreover, people only ask the question `Is there a clock tower in the picture?' on images actually containing a clock tower. As experimented by ~\cite{vqa2}, blindly answering \emph{yes} for the question type \emph{Do you see a} without reading the rest of the question or looking at the associated image results in a VQA accuracy of 87\%. And 2) model deficiency. The predictive capability of the language over images from today's VQA models have been corroborated by ablation studies in ~\cite{analyzing}. Figure~\ref{fig:prior1} and ~\ref{fig:prior2} also validate that these VQA methods suffer from the language prior problem.

Though many studies mentioned the language prior problem, nevertheless, little attention has been paid to develop an evaluation metric to quantitatively measure the extent of language prior effect numerically. Therefore, in this paper, we tentatively propose a metric to effectively measure the VQA models' language prior degree and give full validation of this metric in Section~\ref{experiment_results}. We hope this metric can facilitate the advancement of VQA models and other domains which also suffer from the language prior problem.
\subsection{Definition}
In this subsection, we will give a detailed definition and explanation of the proposed metric - language prior score (dubbed as LP score). We first list the main notations used in the metric.

\textbf{Notations.} Let $\bm{\overline{A}}$ denote all the answer multiset\footnote{Allow for duplicate elements.} in the training set, and $\bm{QT}$ be the question type set. For a question type $qt_j$, $\bm{\overline{A}}_j$ indicates the corresponding answer multiset, which is a subset of $\bm{\overline{A}}$; $\bm{A}_j$ indicates the corresponding answer set, which contains the non-redundant elements in $\bm{\overline{A}}_j$. And $n_j^i$ is the number of answer $a^i$ in $\bm{\overline{A}}_j$. For example, let us assume there is only one question type \emph{how many} - $qt_j$, and $\bm{\overline{A}}$ is \{0, 0, 1, 2, 2, 2, 3, 4, 4, 4\}. Now $\bm{\overline{A}}_j$ should be the same as $\bm{\overline{A}}$, then $\bm{A}_j$ is \{0, 1, 2, 3, 4\}. If $a^i$ is 4, then $n_j^i$ should be 3.

\textbf{Answer Precision per Question Type.} After evaluating the model in the validation set, we can compute the answer precision for each question type. We ignore the case that a predicated answer $a^i$ has not been included in the current answer multiset $A_j$ (i.e., $a^i \notin A_j$)\footnote{If most of the predicted answers do not belong to the answer set of the current question types, it is obviously that a very low accuracy will be obtained. In the experiments, only around 0.1\% answers are ignored for all the baselines. Therefore, ignoring those answers ($a^i \notin A_j$) has negligible effects on the language prior measurement.}.  Otherwise we compute $P_j^i$, which is the precision of the predicted answer $a^i$ under the question type $qt_j$, is computed as:
\begin{equation}\label{equ:precision}
  P^i_j = \frac{TP^i_j}{TP^i_j + FP^i_j},
\end{equation}
where $TP^i_j$ denotes the number of true positive answers, i.e., the predicted answer $a_i$ is the same as the ground truth answer under the question type $qt_j$. And $FP^i_j$ denotes the number of false positive answers, i.e., the predicted answer $a_i$ is not consistent with the groundtruth answer under the question type $qt_j$. For example, if a testing question belongs to the question type $qt_j$ and the predicted answer is $a_i$, and then $TP^i_j + 1$ if the groudtruth answer is $a_i$, otherwise  $FP^i_j + 1$.  Apparently, a larger $P_j^i$ indicates that more questions of this type are correctly answered, and vice versa.

\textbf{Language Prior Score.} Let $LP_j^i$ denote the LP score for the predict answer $a^i$ under the question type $qt_j$. Formally, it is defined as:
\begin{equation}\label{equ:LP}
    LP_j^i = (1-P^i_j) * \sigma(\frac{n_j^i}{|\bm{\overline{A}}_j|}),
\end{equation}
where $\sigma$($\cdot$) refers to a non-linear function (here the sigmoid function is adopted) and $|\bm{\overline{A}}_j|$ is the size of multiset $\bm{\overline{A}}_j$. $(1-P^i_j)$ of Equation~\ref{equ:LP} represents the model deficiency when testing. In extreme cases, if a model performs best as oracle, the $P^i_j$ should be near to 1. And accordingly, $(1-P^i_j)$ should be near to 0. $\sigma(\frac{n_j^i}{|\bm{\overline{A}}_j|})$ represents the proportion of the true answer  $a^i$ of a certain question type $qt_j$ in the whole training set. The reason why we use $\sigma$($\cdot$) for smoothing this half is the proportion of different answers varies largely and we hope sparse answers can also contribute to this metric. We can see that a larger $LP_j^i$ is obtained only when 1) the answers of more questions in the validation set (or testing set) are incorrectly predicted to be $a^i$; and 2) the true answers of more questions are $a^i$ in the training set. In other words, if more answers of a question type in the training data are biased towards $a^i$, and more questions of this type are wrongly answered to $a^i$, (i.e., the language prior problem), the larger LP score will be obtained. Therefore, the measurement considers both the training dataset bias and the model deficiency - the two factors that cause the language prior problem as discussed. Finally, the LP score over the whole validation set can be computed as,
\begin{equation}\label{equ:LP-all}
  LP = \frac{1}{|\bm{QT}|} \sum_{j\in{\bm{QT}}} \frac{1}{|\bm{A}_j|}\sum_{i\in{\bm{A}_j}}LP^i_j,
\end{equation}
where $|\bm{QT}|$ is the size of the whole question type set, and $|\bm{A}_j|$ is the size of the answer set under the question type $qt_j$. We can easily conclude that $LP\in[0, 1]$ and the larger the LP score is, the more language prior is resulted in by the model.

\section{Proposed Regularization Method} \label{model}
\begin{figure}
  \centering
  \includegraphics[width=0.95\linewidth]{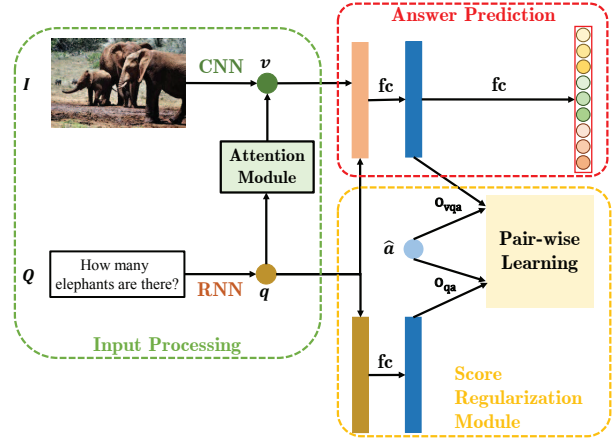}
  \caption{Demonstration of the proposed method for alleviating the language prior problem.}\label{fig:model}
  \vspace{-1em}
\end{figure}


\subsection{Problem Formulation}
The goal of the VQA is to provide an accurate answer $\hat{a}$ to a given textual question $Q$ upon an image or a video $I$. A general approach is to regard the VQA problem as a classification task:
\begin{equation}\label{equ:VQA}
  \hat{a} = \mathop{\arg\max}_{a\in\Omega}p(a|Q, I; \Theta),
\end{equation}
where $\Omega$ denotes the candidate answer set and $\Theta$ denotes the model parameters.

\subsection{Background of VQA Models}
As shown in Figure~\ref{fig:model}, the main framework is composed of three components: \textbf{Input Processing}, \textbf{Answer Prediction} and \textbf{Score Regularization}. The core of our proposed regularization method lies in the \textbf{Score Regularization} part, which will be elaborated in  Section 3.3.
\subsubsection{Input Processing}
There are mainly three parts in the \textbf{Input Processing} component: Image Processing, Question Processing and Attention Module.

\textbf{Image Processing.} The predominant VQA models leverage pre-trained Convolutional Neural Network (CNN) frameworks (e.g., VGG~\cite{vgg} or ResNet~\cite{resnet}) to extract image features $\bm{v}$,
\begin{equation}\label{equ:cnn}
  \bm{v} = \text{CNN}(I).
\end{equation}
As most of the state-of-the-art VQA models adopt the attention mechanism, it is worth mentioning that there are two kinds of image feature extraction. The first one is splitting the image into equal-size regions (e.g., 14$\times$14) and then extracting image features from each region. This will result in a tensor size of $k\times14\times14$ ($k$ is the feature size of each equal-sized image region). The other one is using the region proposal techniques (e.g., Faster R-CNN~\cite{faster}) to extract image features for salient image regions, leading to a tensor size of $k\times n$ ($k$ is the feature size of each image region and $n$ is the proposed salient image region number).

\textbf{Question Processing.} Recurrent Neural Network (RNN, e.g., LSTM~\cite{lstm}) is often used in VQA models to extract question features $\bm{q}$,
\begin{equation}\label{equ:rnn}
  \bm{q} = \text{RNN}(Q).
\end{equation}
More concretely, for a question sentence consisting of $T$ words, its words are fed into the RNN network one by one to obtain their hidden features $\bm{h}$. Usually the last hidden feature $\bm{h}_T$ or all the hidden features (when the attention mechanism is used on each question word) are used to represent this question.

\textbf{Attention Module.}
After the processing of images and questions, a series of image region features \{$\bm{v}_{0}$, $\bm{v}_{1}$, ...\} (14$\times$14 or $n$) and one question feature $\bm{h}_T$ are obtained. Then the VQA models use the question feature to attend on each image region through a multi-layer perceptron (MLP) network or CNN to obtain the attention weights for each image region $\bm{v}_j$:
\begin{equation}\label{equ:attention}
  g_j = \text{ATT}(\bm{h}_T, \bm{v}_j),
\end{equation}
where $g_j$ is normalized through a softmax function. Finally the attended image feature is given by:
\begin{equation}\label{equ:image_att}
  \widetilde{\bm{v}} = \sum \bm{v}_j*g_j.
\end{equation}
\subsubsection{Answer Prediction}
With the attended image feature $\widetilde{\bm{v}}$ and question feature $\bm{h}$, typically, a fusion function (e.g., element-wise addition, element-wise multiplication or concatenation) can be adopted to fuse the question and attended image features. After merging the question and the image features, the VQA models frequently use several linear layers with non-linear activation functions (e.g., ReLU) to make full interactions. Finally the models predict a normalized fixed-length vector and each dimension corresponds to one fixed answer,
\begin{equation}\label{equ:softmax}
  p_{answer} = \text{softmax}(\widetilde{\bm{v}}, \bm{h}).
\end{equation}
The models can be trained by minimizing the log-likelihood loss function, such as:
\begin{equation}\label{equ:loss-answer}
  \mathcal{L}_{answer} = - a_{gt}*\log p_{answer},
\end{equation}
where $a_{gt}$ is the distribution of the ground truth answers.
\subsection{Proposed Regularization Method}
As shown in Figure~\ref{fig:prior1} and ~\ref{fig:prior2}, there are some frequent patterns between question types and answers.  And these patterns are easily captured by the VQA models. As a result, the model will directly give answers based only upon the text questions without referring to the image contents. The VQA then degenerates to a QA problem to some extent.

Based on the above discussion, we would like to achieve that the VQA model can better reason the image contents upon the corresponding questions before predicting the answers, instead of relying on the discovered question-answer patterns to make prediction. To achieve the goal, we design a score regularization module, which adopts a pair-wise learning scheme to make the predicted score obtained from the <image, question, answer> higher than the predicted score obtained from the <question, answer>.

As shown in Figure~\ref{fig:model}, there are two stream inputs to the score regularization module: $\bm{o}_{vqa}$ and $\bm{o}_{qa}$. The former one represents the integration representation of image, question and answer, while the latter one denotes the integration of the question and answer. $\bm{\widehat{a}}$ is the pre-trained word embedding of true answers and it can be fused with other elements (e.g., <image, question> feature or only question feature) to obtain $\bm{o}_{vqa}$ and $\bm{o}_{qa}$. The fusion method includes element-wise addition, multiplication and concatenation. More analysis can be found in Section~\ref{experiment_results}. After this step, the fused features of <image, question, answer> and <question, answer> are used to  predict $s_{vqa}$ and $s_{qa}$,
\begin{align}\label{equ:mlp}
  s_{vqa} = \text{MLP}(\bm{o}_{vqa}), \\
  s_{qa} = \text{MLP}(\bm{o}_{qa}),
\end{align}
where the MLP is leveraged to implement our score regularization module. In order to achieve that questions  with images are better than merely questions for answer prediction, a pair-wise learning method is adopted,
\begin{equation}\label{equ:pair-wise}
  \mathcal{L}_{score} = \max(0, s_{vqa} - s_{qa} + \gamma),
\end{equation}
where $\gamma$ is a relatively small margin. In this way, the backbone models are forced to consider image contents for answering questions, instead of only basing on the frequent patterns between question words and answers.

With the proposed regularization method, the final loss function of the backbone VQA model is a combination of both  the answer prediction loss and the score restriction loss,
\begin{equation}\label{equ:loss}
  \mathcal{L} = \mathcal{L}_{answer} + \beta * \mathcal{L}_{score},
\end{equation}
where $\beta$ is a hyper-parameter balancing these two loss functions. This enables us to train the backbone model with our regularization method in an end-to-end multi-task learning scheme. The default optimization method of the backbone models is kept unchanged to optimize the final loss function.

In Section~\ref{experiment_results}, we will show that different from the methods in~\cite{overcomepriorcvpr, overcomepriornips} which deteriorate the backbone models' performance, our proposed regularization method can boost the backbone models' performance as well as alleviate the language prior problem. 
\section{Experimental Setup} \label{experiment_setup}
We conducted extensive experiments on two datasets to thoroughly justify the effectiveness of our proposed regularization method as well as the feasibility of our proposed evaluation metric. In particular, our experiments mainly answer the following research questions:
\begin{itemize}[align=left,style=nextline,leftmargin=*,labelsep=\parindent,font=\normalfont]
\item \textbf{RQ1}: Can our proposed regularization method outperform the state-of-the-art VQA methods?
\item \textbf{RQ2}: Is the proposed evaluation metric (i.e., LP score) feasible for measuring the extent language prior effect?
\item \textbf{RQ3}: Is the proposed regularization helpful for boosting the answering accuracy as well as alleviating the language prior problem?
\item \textbf{RQ4}: Can backbone models with our proposed regularization method better understand images than those without?
\end{itemize}
\subsection{Datasets}
We testified our proposed method on VQA 1.0~\cite{vqa1} and VQA 2.0~\cite{vqa2} datasets. Both  datasets consist of real images from MSCOCO~\cite{coco} and abstract cartoon scenes. For each image, three different questions are given by Amazon Mechanical Turk (AMT) workers, with ten answers per question. The answers are divided into three categories: \emph{yes/no}, \emph{number} and \emph{other}. Besides, both datasets are split into training, validation and testing (or test-std) splits. The ground truth answers are only available for the first two splits.


%

\subsection{Evaluation Metric}

\textbf{Accuracy.} We adopt the standard accuracy metric for evaluation~\cite{vqa1, vqa2}. Given an image and a corresponding question, for a predicted answer $a$, the accuracy is computed as:
\begin{equation}\label{equ:accuracy}
  Acc_a = \text{min}(1, \frac{\text{\#humans that provide that answer $a$}}{3}).
\end{equation}
Note that each question is answered by ten participants, this metric takes the disagreement in human answers into considerations. The reported results are the averaged accuracy over all questions.

\textbf{LP Score.} As the ground truth answers are not published for the testing set data, we only compute the LP score on the validation set. The computation of LP score is elaborated in Equation~\ref{equ:LP-all}.

\begin{table*}[htbp]
  \caption{Performance of accuracy comparisons between the proposed method and baselines over the VQA 1.0 dataset. The best performance in current splits is highlighted in bold.}\label{tab:baseline1}
  \vspace{-1em}
  \begin{tabular}{lcccccccccccc}
    \toprule[1.2pt]
    \multirow{2}{*}{Method}                 & \multicolumn{4}{c}{Validation}   & \multicolumn{4}{c}{Test-dev}   & \multicolumn{4}{c}{Test-std}  \\
                                            \cmidrule(lr){2-5}                 \cmidrule(lr){6-9}               \cmidrule(lr){10-13}
                                            & Y/N   & Num.  & Other & All      & Y/N   & Num.  & Other & All    & Y/N   & Num.  & Other & All \\
    \midrule
    Question-only~\cite{vqa1}               & 77.86 & 30.24 & 27.61 & 46.75    & 78.20 & 35.68 & 26.59 & 48.76  & 78.12 & 34.94 & 26.99 & 48.89\\
    HieCoAttn~\cite{hierarchical}           & 79.6  & 35.0  & 45.7  & 57.0     & 79.7  & 38.7  & 51.7  & 61.8   &   -   & -     & -     & 62.1 \\
    SAN~\cite{san}                          & 78.6  & 41.8  & 46.4  & 57.6     & 79.30 & 36.60 & 46.10 & 58.70  & 79.11 & 36.41 & 46.42 & 58.85\\
    NMN~\cite{nmn}                          & 80.44 & 34.03 & 40.66 & 54.72    & 81.2  & 38.0  & 44.0  & 58.6   & -     & -     & -     & 58.7 \\
    Strong-baseline~\cite{strongbaseline}   & 82.31 & 35.77 & 51.67 & 61.10    & 82.2  & \bf{39.1}  & 55.2  & 64.5   & 82.0  & 39.1  & \bf{55.2}  & 64.6 \\
    Ask-me-anything~\cite{ask}              & -   & -     & -     & 55.96      & 81.01 & 38.42 & 45.23 & 59.17  & 81.07 & 37.12 & 45.83 & 59.44\\
    SMem~\cite{smem}                        & -     & -     & -     & -        & 80.87 & 37.32 & 43.12 & 57.99  & 80.80 & 37.53 & 43.48 & 58.24\\
    \midrule
    SAN-GVQA~\cite{overcomepriorcvpr}       & 76.90 & -     & -     & 51.12    & -     & -     & -     & -      & -     & -     & -     & -    \\
    SAN+Q-Adv+DoE~\cite{overcomepriornips}  & 71.06 & 32.59 & 42.91 & 52.15    & -     & -     & -     & -      & -     & -     & -     & -    \\
    \midrule
    Ours (Strong-baseline-SR)               & \bf{82.51} & \bf{35.80} & \bf{51.68} & \bf{61.30}
                                            & \bf{83.10} & 39.05 & \bf{55.9} & \bf{65.15}  & \bf{83.2} & \bf{39.14} & 55.12 & \bf{65.28}\\
    \bottomrule[1.2pt]
  \end{tabular}
\end{table*}

\begin{table*}[htbp]
  \caption{Performance of accuracy comparisons between the proposed method and baselines over the VQA 2.0 dataset. The best performance in current splits is highlighted in bold.}\label{tab:baseline2}
  \vspace{-1em}
  \begin{tabular}{lcccccccccccc}
    \toprule[1.2pt]
    \multirow{2}{*}{Method}                     & \multicolumn{4}{c}{Validation}   & \multicolumn{4}{c}{Test-dev}   & \multicolumn{4}{c}{Test-std}  \\
                                                \cmidrule(lr){2-5}                 \cmidrule(lr){6-9}               \cmidrule(lr){10-13}
                                                & Y/N   & Num.  & Other & All      & Y/N   & Num.  & Other & All    & Y/N   & Num.  & Other & All \\
    \midrule
    Question-only~\cite{vqa1, vqa2}             & 67.90 & 30.48 & 26.49 & 42.57    & 67.17 & 31.41 & 27.36 & 44.22  & 67.01 & 31.55 & 27.37 & 44.26\\
    Up-Down~\cite{updown}                       & 80.3  & 42.8  & 55.8  & 63.2     & 81.82 & 44.21 & 56.05 & 65.32  & 82.20 & 43.90 & 56.26 & 65.67\\
    DCN~\cite{dcn}                              &   -   & -     & -     & 62.94    & 83.51 & 46.61 & 57.26 & 66.87  & 83.85 & 47.19 & 56.95 & 67.04\\
    DA-NTN~\cite{ntn}                           & \bf{83.09} & 44.88 & 55.71 & 64.58    & \bf{84.29} & 47.14 & 57.92 & 67.56  & \bf{84.60} & 47.13 & 58.20 & 67.94\\
    Counter~\cite{counter}                      & 81.81 & \bf{49.22} & \bf{56.96} & 65.28  & 83.14 & 51.62 & \bf{58.97} & 68.09  & 83.56 & 51.39 & 59.11 & 68.41\\
    \midrule
    SAN-GVQA~\cite{overcomepriorcvpr}           & 72.03 & -     & -     & 48.24    & -     & -     & -     & -      & -     & -     & -     & -    \\
    SAN+Q-Adv+DoE~\cite{overcomepriornips}      & 69.98 & 39.33 & 47.63 & 52.31    & -     & -     & -     & -      & -     & -     & -     & -    \\
    Up-down+Q-Adv+DoE~\cite{overcomepriornips}  & 79.84 & 42.35 & 55.16 & 62.75    & -     & -     & -     & -      & -     & -     & -     & -    \\
    \midrule
    Ours (Up-down-SR)                           & 80.91 & 43.2  & 55.03 & 63.68    & 81.86 & 44.12 & 56.20 & 66.35  & 82.98 & 43.97 & 56.96 & 66.58\\
    Ours (Counter-SR)                           & 82.48 & 49.02 & 56.88 & \bf{65.29}    & 83.67 & \bf{51.63} & 58.57 & \bf{68.12}
                                                & 83.87 & \bf{51.60} & \bf{59.16} & \bf{68.43}\\
    \bottomrule[1.2pt]
  \end{tabular}
  \vspace{-1em}
\end{table*}

\subsection{Compared Baselines}
We added our regularization method into the following three state-of-the-art baselines. The first one is from the VQA 1.0 dataset, while the last two are from the VQA 2.0 one.
\begin{itemize}[align=left,style=nextline,leftmargin=*,labelsep=\parindent,font=\normalfont]
\item \textbf{Strong-baseline}~\cite{strongbaseline} leverages two stacked ConvNets to obtain the final attention weights for each equal-sized image region. After that it fuses the attentive image feature with the question feature through the vector addition approach.

\item \textbf{Up-Down}~\cite{updown} utilizes a \emph{top-down} mechanism to determine attention weights from \emph{bottom-up } image features (object level and other salient image regions).

\item \textbf{Counter}~\cite{counter} is an upgraded version of the Strong-baseline, introducing a counting module to enable robust counting from object proposals.
\end{itemize}

\subsection{Implementation Details}
We kept most of the setting of backbone models unchanged, including batch size, optimization method, number of non-linear layers. For all the three backbone models, the trade-off parameter $\beta$ was tuned in the range [0.001, 0.01, 0.1, 1, 10, 100]; the margin $\gamma$ was tuned in [0.0, 1.0] with a step size 0.1; and the number of MLP in our score regularization module is fixed to 2; a dropout layer is added between the two layers with a dropout rate 0.5.

\section{Experimental Results} \label{experiment_results}
\subsection{Performance of Accuracy Comparison (RQ1)}
Table~\ref{tab:baseline1} and Table~\ref{tab:baseline2} summarize the accuracy comparison results between our proposed score regularization method with baselines from two groups: traditional VQA models (e.g., Question-only~\cite{vqa1}, NMN~\cite{nmn}, DCN~\cite{dcn}) and VQA models designed to alleviate the language prior problem (i.e., SAN-GVQA~\cite{overcomepriorcvpr} and SAN+Q-Adv+DoE~\cite{overcomepriornips}). The answers are divided into three categories: \emph{Y/N}, \emph{Num.} and \emph{Other}. And the split \emph{All} represents the overall accuracy. Besides, Strong-baseline-SR, Up-down-SR and Counter-SR\footnote{The fusion method between <question, image> and answer features is element-wise multiplication. More analysis can be found in Section 5.3.} are backbone models Strong-baseline, Up-down and Counter with our regularization method, respectively.

For VQA models from the second group, the observation from Table~\ref{tab:baseline1} and Table~\ref{tab:baseline2} is that these VQA models all deteriorate the corresponding backbone models's performance. For example, the overall accuracy deterioration on the validation set of SAN-GVQA over backbone model SAN is 5.88\% on Table~\ref{tab:baseline1}, and Up-down+Q-Adv+DoE over backbone model Up-down is 0.45\% on Table~\ref{tab:baseline2}.   Compared with the models in this group, the final models with our score regularization module (i.e., Strong-baseline-SR, Up-down-SR and Counter-SR) can outperform these  baselines with a large margin on both the VQA 1.0 and 2.0 datasets. For example, on the VQA 1.0 dataset, the absolute improvement of Strong-baseline-SR over SAN+Q-Adv+DoE on Validation All is 9.15\%; on the VQA 2.0 dataset, Counter-SR over Up-down+Q-Adv+DoE is 2.74\%.

Note that the methods in the second group are carefully designed to alleviate the language prior problem, however, there is no evidence in their reports to validate their effects. That means although those methods can indeed alleviate the language prior problem, we still do not know to what extent they can achieve\footnote{The codes of SAN-GVQA~\cite{overcomepriorcvpr},  SAN+Q-Adv+DoE and Up-down+Q-Adv+DoE~\cite{overcomepriornips} are not available, and it is hard for us to replicate their results due to the complicate parameter tuning in the SAN model, therefore, we cannot get the LP scores for those models.}. In the next, we analyze the feasibility of our proposed metric LP score and  use it to measure the extent of language prior effect of the model with and without our regularization model.


\subsection{Feasibility of the Proposed Metric (RQ2)}
\subsubsection{Case Analysis}
We chose two question types \emph{how many} and \emph{what animal} to analyze the feasibility of the proposed LP score metric. The answer distribution in the training set of question type \emph{what animal} is much more uniform than that of \emph{how many}. Note that the \emph{Question-only} method answers the questions merely based on the question features without reasoning images which will arise the language prior problem certainly. From Table~\ref{tab:question-type} we could see that the LP scores of the state-of-the-art approaches are lower than that of the \emph{Question-only} ones, which is consistent with that the language prior problem affects smaller on the former ones than the latter ones. Moreover, for question type \emph{how many}, the LP scores of state-of-the-art methods and the regularized methods\footnote{The fusion method is element-wise multiplication.} on both the VQA 1.0 and VQA 2.0 datasets are just slightly better than the \emph{Question-only} one, respectively. In contrast, for the more uniform answer distribution of question type \emph{what animal}, there is a large margin between the state-of-the-art models and the Question-only model. This indicates that for the state-of-the-art VQA models, the language prior effect of these question types with less uniform answer distributions is higher than those with more uniform answer distributions. Based on the analysis, we can deduce that our proposed metric is capable of measuring the language prior effect.

\subsubsection{Overall Analysis}
\begin{figure}
  \centering
  \includegraphics[width=0.95\linewidth]{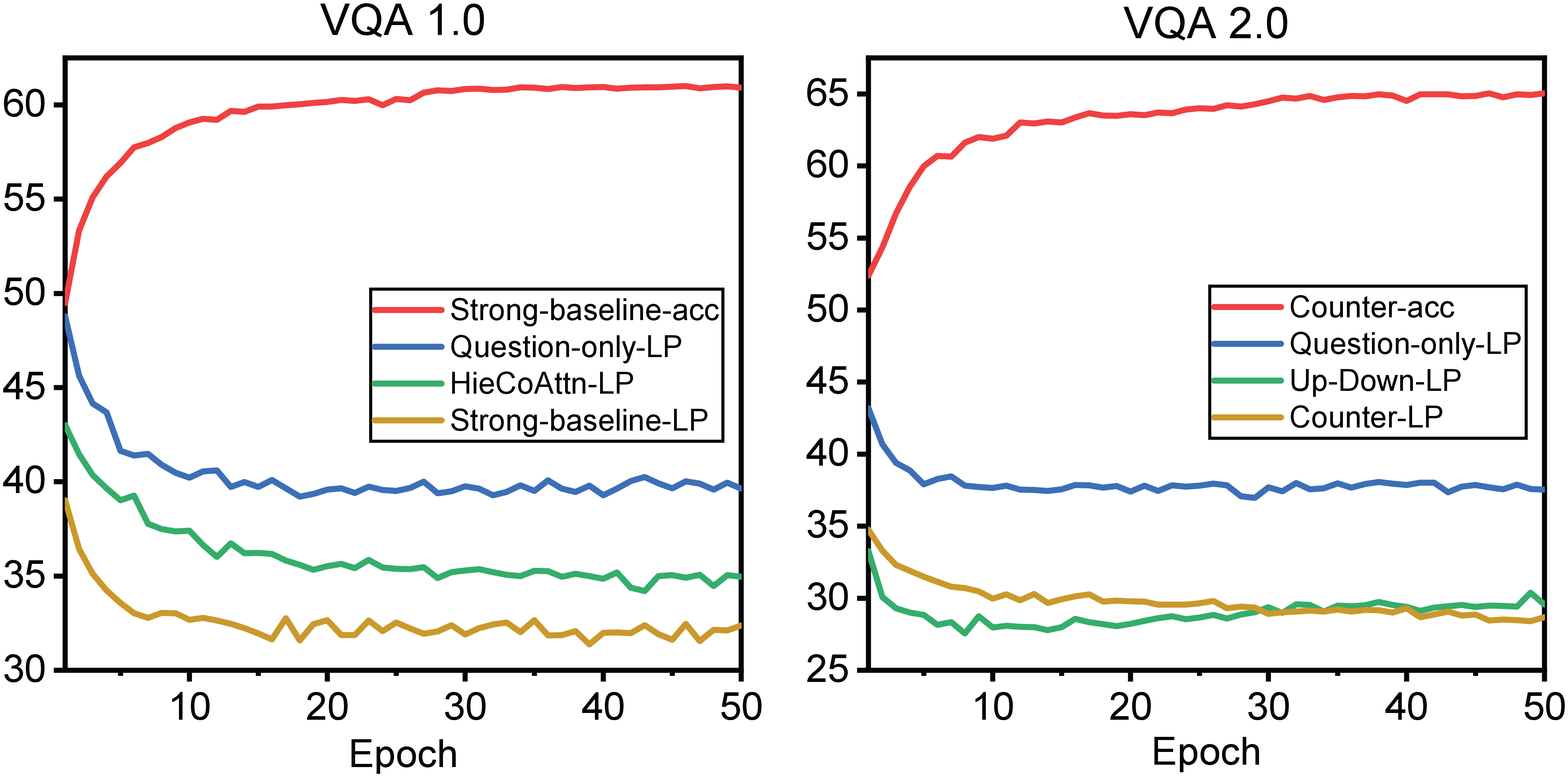}
  \caption{The convergence illustration of LP scores and accuracy over several baselines.}\label{fig:lp}
  \vspace{-2em}
\end{figure}

Figure~\ref{fig:lp} shows how the LP score and accuracy changes with the increase of training epochs. The red line shows the accuracy of one typical baseline model, while the other three on each sub-figure show the LP scores of three baselines.  At the very beginning, the VQA models answer questions mainly based on the learned language prior, which results in a higher LP score in the first few training epochs. With more iterations on the training set, the LP scores begins to drop and the accuracy begins to rise. This denotes that the current VQA models learn to weaken the influence of language prior problem so that the overall accuracy can obtain improvement. If more language prior can be alleviated or overcome, there should be accuracy improvement instead of accuracy degradation. Therefore, it is promising to study and alleviate the language prior problem.

\begin{table}
  \begin{tabular}{rcccc}
    \toprule
    \small
    \multirow{2}{*}{Methods}    & \multicolumn{2}{c}{How many}  & \multicolumn{2}{c}{What animal}\\
                                \cmidrule(lr){2-5}
                                & VQA 1.0       & VQA 2.0       & VQA 1.0       & VQA 2.0           \\
    \midrule
    Question-only               & 50.37         & 49.80         & 54.49         & 53.55             \\
    Strong-baseline             & 49.89         & -             & 33.84         & -                  \\
    Strong-baseline-SR          & 49.81         & -             & 33.85         & -                  \\
    Up-down                     & -             & 48.01         & -             & 33.81              \\
    Up-down-SR                  & -             & 47.90         & -             & 33.69              \\
    Counter                     & -             & 46.30         & -             & 31.09              \\
    Counter-SR                  & -             & 46.26         & -             & 31.04              \\
    \bottomrule
  \end{tabular}
  \caption{LP scores of four baselines and three regularized methods on two typical question types.}\label{tab:question-type}
  \vspace{-2em}
\end{table}

\subsection{Effect of the Proposed Method (RQ3)}
Table~\ref{tab:score1} and Table~\ref{tab:score2} show the influence of our score regularization module over three baselines on the VQA 1.0 and VQA 2.0 datasets, respectively. The second and the last column report the accuracy metric, while the third column reports the LP score metric. To be more specific, \emph{mul} represents that the element-wise multiplication is used for the feature fusion of <image, question, answer> or <question, answer>  in score regularization module Equation~\ref{equ:mlp}, \emph{add} denotes element-wise addition, while \emph{con} represents concatenation. We could observe that different from SAN-GVQA~\cite{overcomepriorcvpr} and SAN+Q-Adv+DoE~\cite{overcomepriornips} deteriorating the backbone models, our proposed regularization method can achieve comparative performance or boost the backbone accuracy performance (e.g., Up-down-SR (mul) over Up-down is 0.48\%). Moreover, the LP score of the proposed regularization method can also outperform the corresponding backbone models. This demonstrates the advancement of our regularization method over the existing ones that we can alleviate the language prior problem as well as boost the backbone models' performance.
\begin{table}
  \caption{Influence of the score regularization module on the VQA 1.0 dataset.}\label{tab:score1}
  \vspace{-1em}
  \begin{tabular}{lccc}
    \toprule[1.2pt]
    Method                          & Valid-All     & Valid-LP      & Test-dev-All \\

    \midrule
    Strong-baseline                 & 61.10         & 31.54         & 64.5       \\
    Strong-baseline-SR (mul)        & \bf{61.30}    & 31.36         & 65.11      \\
    Strong-baseline-SR (add)        & 61.19         & \bf{31.33}    & 64.88      \\
    Strong-baseline-SR (con)        & 61.13         & 31.38         & \bf{65.15} \\
    \bottomrule[1.2pt]
  \end{tabular}
  \vspace{-1em}
\end{table}

\begin{table}
  \caption{Influence of the score regularization module on the VQA 2.0 dataset.}\label{tab:score2}
  \vspace{-1em}
  \begin{tabular}{lccc}
    \toprule[1.2pt]
    Method                          & Valid-All     & Valid-LP      & Test-dev-All \\
    \midrule
    Up-down                         & 63.20         & 29.71         & 65.32      \\
    Up-down-SR (mul)                & \bf{63.68}    & 29.44         & 66.35      \\
    Up-down-SR (add)                & 63.53         & \bf{29.43}    & 66.25      \\
    Up-down-SR (con)                & 63.55         & 29.50         & \bf{66.46} \\
    \midrule
    Counter                         & 65.28         & 29.74         & 68.09      \\
    Counter-SR (mul)                & \bf{65.29}    & \bf{29.67}    & \bf{68.12}      \\
    Counter-SR (add)                & 65.03         & 29.84         & 67.88      \\
    Counter-SR (con)                & 65.01         & 29.88         & 67.86      \\
    \bottomrule[1.2pt]
  \end{tabular}
  \vspace{-1em}
\end{table}

\subsection{Visualization of Attention Kernels (RQ4)}
\begin{figure*}
  \centering
  \includegraphics[width=0.95\linewidth]{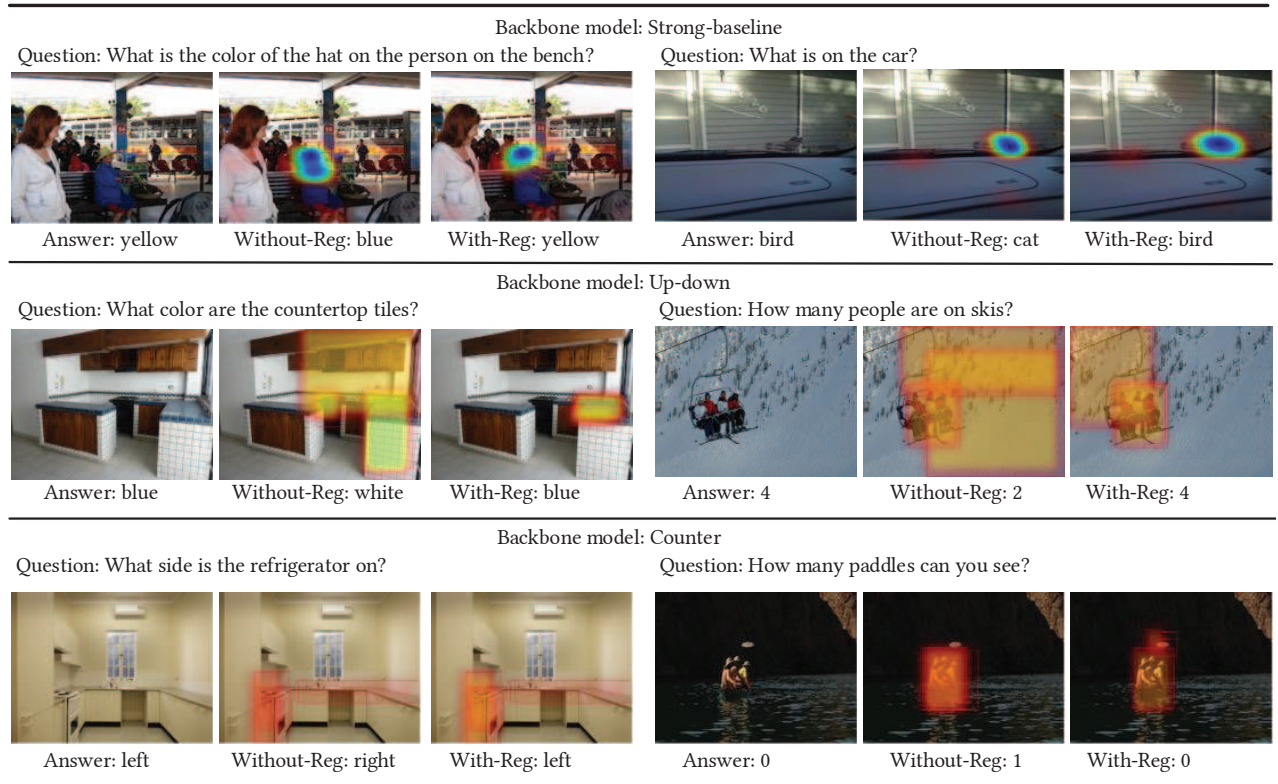}
  \vspace{-1em}
  \caption{Visualization of three backbone models with and without the proposed regularization method.}\label{fig:visulization}
  \vspace{-1em}
\end{figure*}

As the attention module becomes an indispensable part of current VQA models, we visualized some examples of the attention kernels from these backbone models with and without our score regularization module. We mainly listed the questions which belong to the question type \emph{how many} and \emph{what color}, other more uniform question type samples are also analyzed in Figure~\ref{fig:visulization}, e.g., \emph{what is}. There are three rows of six examples, each row illustrates two samples from one backbone model with and without regularization, where the backbone models without regularization predicted incorrectly and backbone models with regularization predicted correctly. From the figure, we can see the failure cases of VQA methods without regularization can be grouped into two categories: 1) attending to wrong regions and predicting answers incorrectly and 2) attending to correct regions but predicting answers wrongly.

Both samples from the second row belong to the first category. For instance, the first example is about the color of the countertop tiles, and the backbone model Up-down without regularization focuses on the closet and the white tile while the true region should be the tile on top of the countertop. As illustrated in Figure~\ref{fig:prior1}, the number of the wrong answer of \emph{white} is much larger than that of the true answer \emph{blue} in the VQA 1.0 training set, which leads to the language prior problem here. The second example from backbone model Up-down shares the same problem. In contrast, examples falling into the second category attended to correct image regions but predicting answers incorrectly. For instance, the second example from Strong-baseline is similar to the image classification task, where the true answer should be a bird instead of a cat. And the second example of backbone model Counter belongs to \emph{how many} question type. Since the number of the answer \emph{1} is much more than that of the answer \emph{0} under this question type in the VQA 2.0 dataset, this example also testified that the language prior learned by the model Counter causes a wrong prediction and it can be corrected by our regularization method (as shown in the result of Counter-SR).

\section{Related Work} \label{related_work}
\subsection{Visual Question Answering}
Traditional text-based QA~\cite{textqa, chen2019driven, qa} has been long recognized as a challenging information retrieval task. Derived from it, other QA systems like community QA (CQA)~\cite{communityqa}, multimedia QA~\cite{multimediaqa}and visual QA (VQA)~\cite{vqa1, vqa2, mfb} have attracted researchers' interest in recent years. We mainly recap the related studies of VQA in this subsection.


VQA witnesses a renewed excitement in multi-discipline AI research problems due to the development of CV, NLP and IR. Generally speaking, the existing VQA methods can be classified into four categories~\cite{survey}: \emph{Joint Embedding}, \emph{Attention Mechanism-based}, \emph{Compositional} and \emph{Knowledge Base-enhanced}. However, the language prior problem is observed across the existing VQA models~\cite{vqa1, ask, san}. It is impossible to distinguish an answer arising because of the image reasoning and one selected because it occurs frequently in the training set. In the view of amending biased datasets, researcher in~\cite{yinyang} added <image, question, answer> triplets by compositing another visually similar image but with an opposite answer to a binary question for VQA 1.0 abstract scenes. Similarly, authors in~\cite{vqa2} added triplets based on all varieties of questions for VQA 1.0 real images. Instead of supplementing biased datasets, authors in ~\cite{clevr} designed a diagnostic 3D shape to balance answer distribution for each question type from scratch. Different from the above ones, methods in ~\cite{overcomepriorcvpr, overcomepriornips} aim to force VQA models to better understand the images. The authors built their models on SAN~\cite{san} with restrictions to prevent the model from exploiting language correlations in the training data.


It is worth emphasizing that the previous methods solve the language prior problem either by introducing other bias into existing datasets~\cite{yinyang, vqa2} or degrading performance over the backbone models~\cite{overcomepriorcvpr, overcomepriornips}. In this paper, we proposed a regularization method for several publicly released state-of-the-art attentional VQA models. In addition to the capability of alleviating the language prior problem, better accuracy is observed for the VQA models with our regularization module than the corresponding ones without.

\subsection{Deep Multimodal Fusion}
In this work, we fuse the multi-modality features - image, question, and answer in the proposed regularization module. Here we briefly review related works in this direction. There is a large amount of studies on integrating multimodal data sources in deep neural networks, including recommendation~\cite{attmm, mmalfm, poimm}, multimodal retrieval~\cite{mmre, search}, and user profiling~\cite{user}, image captioning~\cite{captionsig, caption}. The flexibility of deep architecture advances the implementation of multimodal fusion either as feature-level fusion or decision-level fusion~\cite{mmsurvey}.

Methods in the feature-level fusion group transform the raw inputs from multiple paths into separate intermediate representations, followed by a shared representation layer to merge them. For instance, Chen et al.~\cite{attmm} proposed a two-level attention mechanism to fuse the features from component-level and item-level for multimedia recommendation. Farnadi et al.~\cite{user} utilized a shared representation between different modalities to arrive at more accurate user profiles. Some efforts~\cite{captionsig, caption} in image captioning merged previous word representations and image features to produce the next word.

By contrast, decision-level fusion refers to the aggregation of decisions from multiple classifiers, each trained on separate modalities. These fusion rules could be max-fusion, averaged-fusion, Bayes’ rule based, or even learned using a meta-classifier~\cite{mmsurvey}. For example, the work in~\cite{decision1} presents a two-stream CNN (i.e., Spatial stream ConvNet and Temporal stream Convnet) and then combines them with a \emph{class score fusion} approach for action recognition in videos. In order to achieve simultaneous gesture segmentation and recognition, authors in~\cite{decision2} integrated the emission probabilities estimated from  different inputs (i.e, skeleton joint information, depth and RGB images) as a simple linear combination.


\section{Conclusion} \label{conclusion}
The language prior problem severely hinders the advancement of VQA. In this paper, we target this problem and make contributions from two perspectives. Firstly, we propose an evaluation metric called LP score to measure the extent of language prior effect. The evaluation metric can quantitatively measure the extent of language prior effect of different VQA models and thus can facilitate the development of related techniques. Secondly, we design a score regularization module, which is flexible to be integrated into various current VQA models. The proposed regularizer can effectively make the VQA models better reason images upon questions before result prediction, and thus can alleviate the language prior problem as well as improve the answer accuracy. Extensive experiments have been conducted on two widely used VQA datasets to validate the feasibility of the proposed metric and the effectiveness of the designed regularization method. We hope this metric can be used to compare the VQA models on alleviating the language prior problem in the future. Besides, we would like to further extend our regularizer based on the non-uniform granularity of different question types and explore its effectiveness on more diversified VQA models.

\begin{acks}
This work is supported by the National Natural Science Foundation of China, No.: 61772310, No.:61702300, No.:61702302, No.: 61802231, and No.: U1836216; the Project of Thousand Youth Talents 2016; the Tencent AI Lab Rhino-Bird Joint Research Program, No.:JR201805; and  the National Research Foundation, Prime Minister’s Office, Singapore under its Strategic Capability Research Centres Funding Initiative.
\end{acks}

\clearpage

\bibliographystyle{ACM-Reference-Format}
\balance
\bibliography{sigir19}

\end{document}